\documentclass{article}
\usepackage{iclr2027_conference,times}

\usepackage{amsmath,amsfonts,bm}

\def\eqref#1{equation~\ref{#1}}

\def\1{\bm{1}}

\DeclareMathAlphabet{\mathsfit}{\encodingdefault}{\sfdefault}{m}{sl}
\SetMathAlphabet{\mathsfit}{bold}{\encodingdefault}{\sfdefault}{bx}{n}

\usepackage{hyperref}
\usepackage{url}
\usepackage{booktabs}
\usepackage{graphicx}
\usepackage{wrapfig}
\usepackage{amsmath,amssymb}
\usepackage{enumitem}

\graphicspath{{figures/pdf/}}

\newcommand{\method}{DecSteer}
\newcommand{\kvs}{KV-Skill}

\title{Learning to Decide, Not to Reason: Parameter-Efficient  Decision Operators via Low-Rank Activation Steering}

\author{
	Ran Li\textsuperscript{1} \quad
	Lei Chen\textsuperscript{1,2} \\
	\textsuperscript{1}Hong Kong University of Science and Technology  \\
	\textsuperscript{2}Hong Kong University of Science and Technology(Guangzhou)  \\
	\texttt{rlibb@connect.ust.hk} \quad \texttt{leichen@cse.ust.hk}
}

\iclrfinalcopy

\begin{document}

\maketitle

\makeatletter
\def\@oddhead{\small Preprint. Under review.\hfil}
\def\@evenhead{\small Preprint. Under review.\hfil}
\makeatother

\begin{abstract}
Injecting skills into a frozen language model currently costs a million parameters and a reinforcement-learning pipeline. We introduce \method{}, a System-1 decision operator trained by behavior cloning that lowers this cost by roughly two orders of magnitude. The default operator uses 330K parameters to match a 1.33M-parameter operator trained with reinforcement learning, exceeds or achieve comparable performance, while collapsing 3,685-token deliberation into a 6-token decision with no loss in accuracy. A rank-4 variant with 23K parameters, 1/58 of the strongest published skill operator, suffices for SearchQA and near-suffices for LiveMath, where higher rank still helps; the same recipe transfers across five tasks and three backbones, with out-of-distribution gains persisting on LiveMath problems released months after training. The gap to prior work is  trainability, and it is set jointly by initialization and architecture. The $V{=}0$ initialization of prior operators zeroes the gradient of both large factor matrices at the first optimization step, whereas our zero-initialized output projection inside a shared low-rank backbone receives a gradient immediately, which a gradient-flow probe confirms directly. The gain isn't chain-of-thought compression. 23 of 57 LiveMath points beat the base model's best-of-8 sampling, and a logit-lens probe shows the operator amplifies the answer along the model's existing late-layer pathway, not writing it earlier. Gains track the base model's headroom across 13 base--task pairs, and skills compose as approximately linear operators that can be added, interpolated, and hot-swapped at inference time. Code on {\url{https://github.com/rlisml/decisionsteer}}.
\end{abstract}

\section{Introduction}
\label{sec:intro}

\begin{wrapfigure}{r}{0.49\linewidth}
\vspace{-1cm}
\centering
\includegraphics[width=\linewidth]{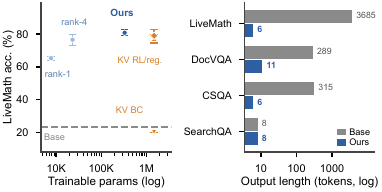}
\caption{A decision does not need a million parameters. \textbf{Left:} LiveMath accuracy vs.\ trainable parameters. A 23K-parameter model trained by behavior cloning matches a 1.33M-parameter one trained with RL, while the same BC fails on the prior one. \textbf{Right:} It collapses median output length from 3,685 to a 6-token decision.}
\label{fig:hero}
\end{wrapfigure}

When a language model already holds the skill a task requires, deliberating over it is wasted compute. The distinction between slow deliberation and fast decision has returned to the center of discussion as reasoning models spend thousands of tokens on problems they could answer directly \citep{wei2022chain,sui2025stop,fang2026thinkless}, and a line of work asks how System-2 reasoning can be compiled into System-1 behavior \citep{yu2024distilling,zelikman2024quietstarlanguagemodelsteach,su2025dualformer}. A recent and striking demonstration of this compilation is \kvs{} \citep{han2026kv}: a small external operator, grafted onto the residual stream of a frozen 4B model and trained with reinforcement learning, lifts LiveMath accuracy from 23.4 to 79.0 and collapses long reasoning into direct answers. This, however, comes with a cost structure: 1.33M trainable parameters, a GRPO rollout server, and minutes-to-hours of RL tuning. That cost structure shapes how skills can be used. If a skill costs a million parameters and an RL pipeline, one trains few skills carefully; if a skill costs minutes on one GPU, one can afford a library of them.

This paper asks what a skill actually costs. Our answer is that the published cost is about fifty times too high, and that the reinforcement learning is unnecessary. \method{} is a low-rank residual operator, applied at eight decoder depths in the spirit of activation steering \citep{turner2023steering,li2023inference,wu2024reft,adila2026weightupdatesactivationshifts}, with three deliberate choices: a single shared rank-$r$ backbone normalized by RMS and gated per depth, a zero-initialized output projection, and behavior cloning on answer tokens only. The default configuration uses 330K parameters and 12 minutes of single-GPU BC; the rank-4 configuration uses 23K. Against \kvs{} under identical data, splits, and evaluators, \method{} matches or exceeds every published variant while training by behavior cloning alone. On LiveMath it reaches 80.9 (RL baseline 79.0); on SearchQA 81.1 (80.1); the 6-token decision it emits replaces a 3,685-token median reasoning chain and runs 36$\times$ faster. The skill also holds out of distribution. A bundle trained only on problems released through February keeps its full gain on new problems and stays ahead of the officially released RL artifact.

Three observations convinced us this is a property of skills rather than a favorable engineering accident, and they organize the paper.
\textbf{Why has this not been found before?} \kvs{} itself ran the obvious control. It trained its own 1.33M operator with the same BC signal and the same 400 gold examples, and the result sat at the base model (68.4 on SearchQA, 20.2 on LiveMath, versus base 68.1 and 22.4). It is tempting to conclude from this that BC is too weak a signal for skills. What separates the two cases is the identity anchor together with the architecture built around it. \kvs{} anchors its operator at the base model by zero-initializing a value matrix $V$ that appears in the injection path, which zeroes the gradient of \emph{both} large factor matrices at the first step; only a small middle matrix moves early. Our $up{=}0$ anchor zeroes the output projection instead, so the large output matrix receives a gradient of size $\Theta(\sqrt{d_s})$ immediately. A gradient probe confirms the two paths step by step (Section~\ref{sec:mechanism}), and a rank scan shows the resulting knee. 23K parameters suffice, and on SearchQA even a rank-1 operator (7.7K parameters, 1/173 of \kvs{}) exceeds the published RL result.
\textbf{Is the gain just compressed reasoning?} No. A chain of controls on LiveMath decomposes the 57-point gain. 15 points come from stopping meandering reasoning (a prompting control reaches this for free), 19 points from answers that sampling can touch but greedy decoding cannot, and at least 23 points lie beyond what the base model reaches with best-of-8 sampling, on problems where all 8 samples fail (Section~\ref{sec:mechanism}). A logit-lens probe localizes the mechanism: the operator does not move answer information into earlier layers; it amplifies the answer along the model's existing late-layer output pathway, by 3.4$\times$ to 6.4$\times$, and the amplification grows as the base weakens.
\textbf{When should one steer?} The gain tracks the base model's headroom consistently across the 13 base--task pairs we measured, spanning three backbones (Section~\ref{sec:when}). This empirical trend turns flat results into expectations: on tasks where the 4B base is saturated, every skill arm lands within a point of the base, and the same operator applied to weaker backbones recovers gains of +2.4 to +10.0 points on those same tasks. Skills also behave as approximately linear operators: adding, interpolating, and reordering them is safe across all 18 tested cells, with one clearly marked forbidden zone (negative coefficients on mode-switching skills), so a trained library can be hot-swapped per request at inference time.

\textbf{Contributions:}
\noindent
\begin{itemize}[leftmargin=*,labelindent=0pt,labelsep=0.5em,itemsep=0pt,topsep=1pt,parsep=0pt]
\item \method{}, a System-1 decision operator that matches a 1.33M-parameter RL-trained skill operator with 23K--330K parameters, 12 minutes of BC, and a 6-token output, across five tasks and three backbones, including out-of-distribution generalization tasks (Section~\ref{sec:results}).
\item A trainability analysis with a direct gradient probe: trainability is set jointly by the initialization and the surrounding architecture, $V{=}0$ anchoring freezes both large factor matrices at the start while our zero-initialized output projection stays on the optimization path, which explains why prior BC controls failed and where the rank threshold comes from (Section~\ref{sec:mechanism}).
\item A gain attribution showing the operator injects answers beyond the base model's best-of-8 reach, with logit-lens localizing late-layer amplification, not early injection  (Section~\ref{sec:mechanism}).
\item An empirical headroom trend that anticipates when steering pays, and an operator algebra (add, interpolate, reorder, hot-swap) with an explicit boundary (Section~\ref{sec:when}).
\end{itemize}

\section{Background: skill operators and identity anchoring}
\label{sec:background}

\textbf{Skill operators.} \citet{han2026kv} train an \emph{external factorized operator} on the residual stream of a frozen LM. At $D$ decoder depths, the block output $h \in \mathbb{R}^{d_m}$ is perturbed by
\begin{equation}
x = \mathrm{RMS}(h), \quad
q = \mathrm{normalize}(xA^\top), \quad
a = qK, \quad
\Delta h = g \odot (aV^\top)B^\top, \quad
h' = h + \Delta h,
\label{eq:kvskill}
\end{equation}
where $A \in \mathbb{R}^{d_s \times d_m}$ and $B \in \mathbb{R}^{d_m \times d_s}$ are shared down/up projections, $K, V \in \mathbb{R}^{d_s \times r}$ implement a slot-retrieval bottleneck of rank $r$, and $g$ is an input-dependent gate. With $d_s{=}256$, $r{=}32$ this is 1,329,668 parameters, trained with GRPO over rollout batches. The operator is one point in a broader effort to externalize competence into portable objects, from textual skill files and their distillation \citep{tang2026knowledge,lu2026skill0,bai2026skillzip,li2026skillsinjector,yang2026skillopt,tao2026softskill,wang2026skillx} to parameter-space adapters \citep{li2021prefix,lester2021power,liu2022p} and representation finetuning \citep{wu2024reft}.

\textbf{Identity anchoring.} A skill operator must not perturb the model before or without training, so its parameters start at an exact identity. \kvs{} achieves this with $V{=}0$: since $\Delta h$ is proportional to $V$, a zero $V$ guarantees $\Delta h \equiv 0$. The same guarantee is available cheaper. If the \emph{output} projection is zero-initialized, the perturbation vanishes regardless of everything upstream. The two anchors are behaviorally identical at $\theta^0$; Section~\ref{sec:mechanism} shows they differ as optimization starting points, and that the anchor, together with the surrounding architecture, decides whether the large matrices train at all.

\section{\method{}: System-1 decision operators}
\label{sec:method}

\begin{figure}[t]
\centering
\includegraphics[width=0.98\linewidth]{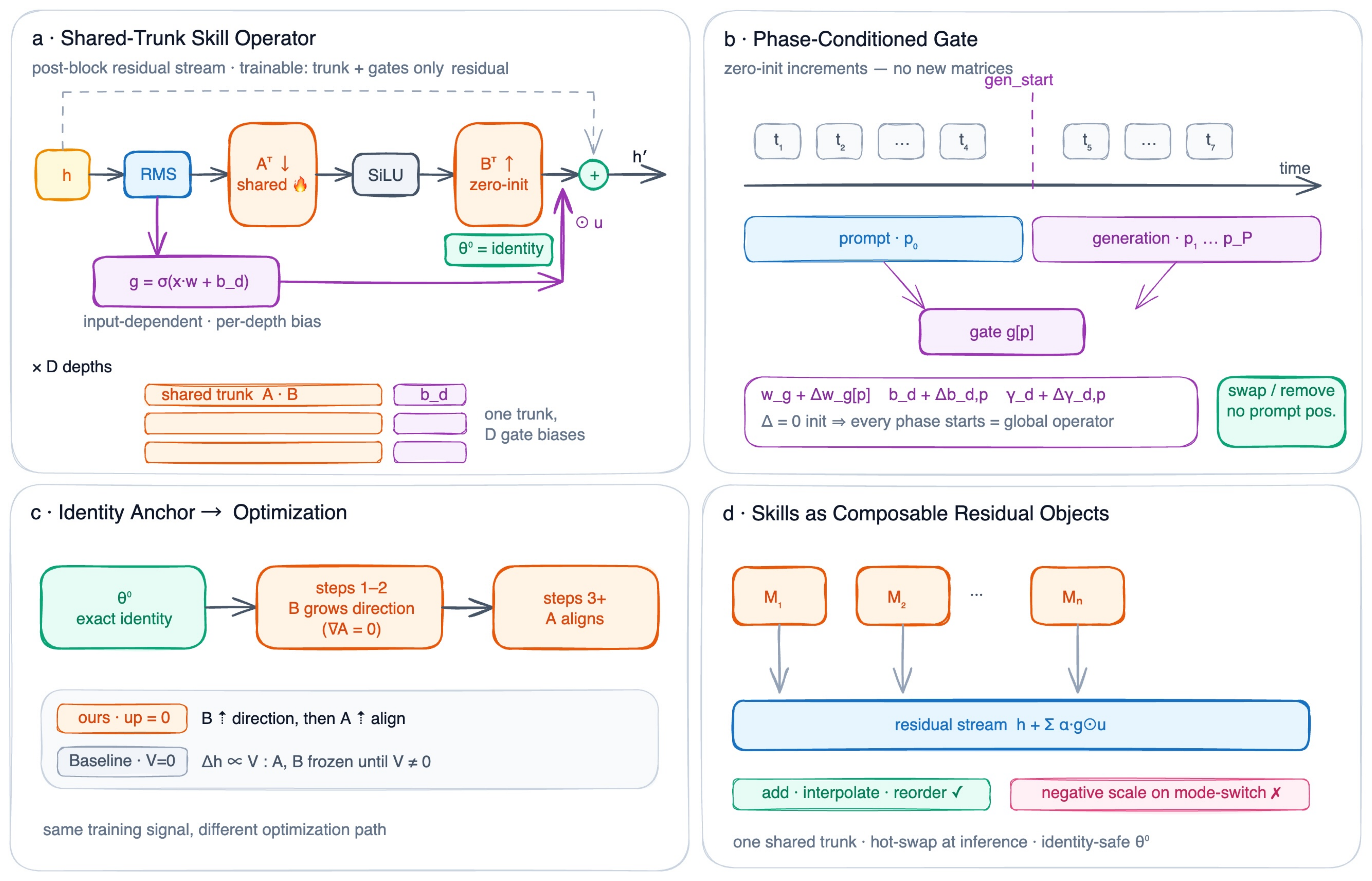}
\caption{Overview. \textbf{a:} A shared low-rank backbone is injected at $D{=}8$ depths with per-depth input-dependent gate $g$; the operator is identity at initialization. \textbf{b:} The gate can be phase-conditioned on prompt vs.\ generation via zero-init increments. \textbf{c:} The identity anchor steers optimization: with $up{=}0$, the output projection gets gradient from step one, while baselines $V{=}0$ keeps both large matrices frozen until $V$ moves. \textbf{d:} Trained skills act as composable residual objects.}
\label{fig:arch}
\end{figure}

\textbf{Operator.} At each of $D{=}8$ depths, \method{} perturbs the block output $h$ by
\begin{equation}
x = \mathrm{RMS}(h), \quad
z = \phi(xA^\top), \quad
u = zB^\top, \quad
g = \sigma(xw_g^\top + b_d), \quad
h' = h + \alpha \, g \odot u,
\label{eq:decsteer}
\end{equation}
with $\phi$ the SiLU nonlinearity and $\sigma$ the sigmoid. The backbone $(A, B, w_g)$ is \emph{shared across depths}; only the gate bias $b_d$ is per depth. $A \in \mathbb{R}^{d_s \times d_m}$ is randomly initialized, $B \equiv 0$ and $w_g \equiv 0$ at $\theta^0$, so $\Delta h \equiv 0$ exactly and the operator starts as a bit-precise identity. RMS normalization decouples the operator from the residual-stream magnitude, which varies by orders of magnitude across depths and is what makes a shared backbone trainable at all; the gate restores depth- and input-specific amplitude that sharing removes. For $d_m{=}2560$, $d_s{=}64$ the parameter count is $2 d_s d_m + d_m + D = 330{,}248$, a quarter of \kvs{}. Depth sharing is where the savings come from: with the same per-depth rank, per-depth independent projections need 2.62M parameters to reach the same score, and adding injection depths beyond eight costs one scalar per depth.

\textbf{Training.} The training signal is behavior cloning with loss computed on answer tokens only, using the same 400 (SearchQA) or 35 (LiveMath) gold examples as \kvs{}'s own BC control. LiveMath trains in 87 steps, 12 minutes, peak 13.9 GiB on a single A100-40GB; no rollouts, no verifier loop, no reference policy. We deliberately keep the signal identical to the published control that failed, so that any difference in outcome is attributable to the operator, not the data.

\textbf{Inference.} The shared backbone is materialized per depth into a single vector expression and injected as a steering vector, so the operator adds no custom serving path and runs at full generation speed. A scalar $\alpha$ exposes an inference-time strength knob; swapping bundles swaps skills per request. Section~\ref{sec:when} returns to what this enables.

\section{Experiments I: decisions are cheap}
\label{sec:results}

\textbf{Setup.} We evaluate on LiveMath \citep{he2026livemathematicianbench}, SearchQA \citep{dunn2017searchqa}, CommonsenseQA \citep{talmor2019commonsenseqa}, OpenBookQA \citep{mihaylov2018can}, and DocVQA \citep{mathew2021docvqa}, with Qwen3.5-4B \citep{yang2025qwen3} as the main backbone and Llama-3.2-3B and Qwen3.5-0.8B for cross-backbone tests. Splits, prompts, decoding (greedy, stop at \texttt{</answer>}), and evaluators are identical to \kvs{}; the Base and baseline rows of Table~\ref{tab:main_results} quote its published scores. Key arms use 3 training seeds and report mean $\pm$ SD; identity anchors, contamination checks, and the strength protocol are documented in Appendix~\ref{app:setup}.

\begin{table*}[t]
	\centering
	\setlength{\tabcolsep}{4pt}
	\caption{Main results. \textbf{Ours} is a low-rank residual operator trained
		by behavior cloning on answer tokens only. KV-Skill, LoRA, Text Skill,
		SkillOpt, and Soft Skill rows quote published numbers; the Base rows are
		their published base scores. Rank rows are trained from scratch with the
		same recipe. In the smaller-backbone blocks, where no published baselines
		exist, the \kvs{} row is our faithful reimplementation trained with the
		identical BC recipe (Appendix~\ref{app:setup}). Ours entries are means over
		3 seeds; a uniform operator strength of 1.0 yields the same conclusions.}
	\label{tab:main_results}
	\begin{tabular}{l c ccccc}
		\toprule
		& & \shortstack{LiveMath\\(Acc)} & \shortstack{SearchQA\\(EM)} & \shortstack{CSQA\\(Acc)} & \shortstack{OpenBookQA\\(Acc)} & \shortstack{DocVQA\\(ANLS)} \\
		\midrule
		\multicolumn{7}{l}{\emph{Qwen3.5-4B}} \\
		Base                    & 0      & 22.4 & 68.1 & 80.7 & 90.6 & 86.9 \\
		Text Skill              & 0      & 23.4 & 67.4 & 78.5 & 89.0 & 88.4 \\
		SkillOpt                & --     & 52.0 & 71.2 & --   & --   & 89.0 \\
		Soft Skill              & --     & 64.5 & 76.4 & --   & --   & 88.2 \\
		LoRA (BC)               & 1.38M  & 19.4 & 69.1 & --   & --   & 93.7 \\
		LoRA (RL)               & 1.38M  & 80.7 & 70.1 & --   & --   & -- \\
		\kvs{} (reg.)           & 1.33M  & 77.2$\pm$2.6 & 79.3$\pm$0.7 & 79.3$\pm$1.2 & 88.7$\pm$0.8 & -- \\
		\kvs{} (RL-scr.)        & 1.33M  & 75.6$\pm$4.1 & 80.0$\pm$3.4 & \textbf{81.1}$\pm$1.6 & \textbf{90.8}$\pm$2.2 & 94.2$\pm$2.9 \\
		\kvs{} (RL-skill)       & 1.33M  & 79.0$\pm$3.9 & 80.1$\pm$3.3 & -- & -- & \textbf{95.0}$\pm$2.6 \\
		\kvs{} (BC)             & 1.33M  & 20.2 & 68.4 & --   & --   & -- \\
		Ours (BC)               & 0.33M  & \textbf{80.9$\pm$1.8} & \textbf{81.1$\pm$1.1} & 80.5$\pm$1.0 & 89.8$\pm$0.2 & 94.8$\pm$0.7 \\
		Ours (rank-1)           & 7.7K   & 65.0$\pm$2.8 & 81.1$\pm$1.3 & -- & -- & -- \\
		Ours (rank-4)           & 23K    & 74.8$\pm$2.6 & \textbf{81.3}$\pm$1.1 & -- & -- & -- \\
		\midrule
		\multicolumn{7}{l}{\emph{Llama-3.2-3B}} \\
		Base                    & 0      & 16.9 & 57.1 & 71.3 & 74.6 & -- \\
		\kvs{} (BC)             & 1.59M  & 66.1$\pm$1.4 & 76.5$\pm$0.1 & 71.7$\pm$0.6 & \textbf{78.9$\pm$0.8} & -- \\
		Ours (BC)               & 0.40M  & \textbf{71.2$\pm$2.0} & \textbf{77.6$\pm$1.3} & \textbf{72.7$\pm$0.9} & 78.2$\pm$0.7 & -- \\
		\midrule
		\multicolumn{7}{l}{\emph{Qwen3.5-0.8B}} \\
		Base                    & 0      & --   & 36.1 & 50.9 & 63.2 & -- \\
		\kvs{} (BC)             & 0.54M  & 62.6$\pm$3.1 & 70.1$\pm$0.7 & 61.0$\pm$1.3 & 66.7$\pm$0.6 & -- \\
		Ours (BC)               & 0.13M  & \textbf{70.7$\pm$3.3} & \textbf{72.9$\pm$0.8} & \textbf{61.1$\pm$1.1} & \textbf{69.8$\pm$1.2} & -- \\
		\bottomrule
	\end{tabular}
\end{table*}

\begin{figure}[t]
\centering
\includegraphics[height=6.0cm]{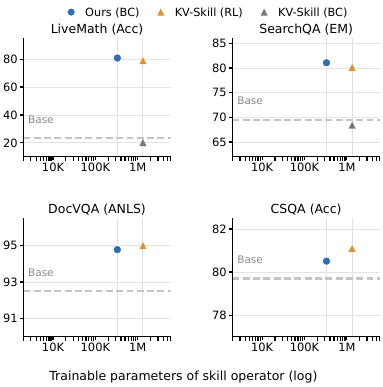}\hspace{0.03\linewidth}%
\includegraphics[height=6.0cm]{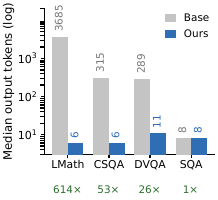}
\caption{\textbf{Left:} accuracy versus trainable parameters per task. \method{} (BC) at 0.33M reaches or exceeds every \kvs{} variant at 1.33M, while \kvs{}'s own BC control (same signal, same examples) stays at the base. \textbf{Right:} median generated tokens, base versus \method{}. LiveMath drops from 3,685 to 6 tokens (36$\times$ wall-clock), DocVQA from 289 to 11 (10.3$\times$), with accuracy unchanged or higher; SearchQA already answers directly.}
\label{fig:pareto}
\end{figure}

\textbf{Main results.} Table~\ref{tab:main_results} and Figure~\ref{fig:pareto} (left) give the headline comparison. With 0.33M parameters and BC only, \method{} reaches 80.9 on LiveMath (best \kvs{} RL: 79.0$\pm$3.9, and with smaller variance), 81.1 on SearchQA (80.1), 94.8 on DocVQA, and holds saturated tasks within a point of the best RL numbers. The BC row of \kvs{} is the instructive one: identical signal, identical examples, 1.33M parameters, and the result does not leave the base (20.2, 68.4). The same signal through our operator trains to 80+. Below 0.33M the story continues rather than degrades: a rank-4 operator (23K, 1/58 of \kvs{}) reaches 81.3$\pm$1.1 on SearchQA and 74.8$\pm$2.6 on LiveMath, within the seed spread of the 1.33M RL baseline, and rank-1 (7.7K) already exceeds every published SearchQA number. On the two smaller backbones, where the published base scores are lower and no RL baselines exist, the identical BC recipe lifts LiveMath by +54 to +60 points and SearchQA by +15 to +37 points over the base, and against a faithful same-protocol reimplementation of the \kvs{} operator it is higher in 6 of 8 cells and never lower, at a quarter of the parameters (Appendix~\ref{app:results}).

\textbf{Training signal.} Under GRPO matched to \kvs{}'s protocol item by item, \method{} matches the published numbers with 1/4.02 of the parameters (Appendix~\ref{app:results}). The instructive cell is GRPO from scratch on LiveMath: it reaches only 71.8, while the same RL started from the BC operator reaches 80.7. On tasks whose target is a single forward mapping, the starting point matters more than the algorithm, which is the same conclusion the trainability analysis of Section~\ref{sec:mechanism} reaches from the gradient side.

\textbf{Inference cost.} The operator learns to decide, and the decision is short. On LiveMath the median output falls from 3,685 tokens of reasoning to a 6-token answer and generation time from 175\,s to 4.8\,s per batch (36$\times$); on DocVQA from 289 to 11 tokens (10.3$\times$ wall-clock); accuracy is unchanged or higher throughout, and on SearchQA, where the base already answers briefly, the direct-answer rate still rises from 95.4\% to 98.8\%. The compression is a property of the training objective (Section~\ref{sec:mechanism}), and it arrives on top of the accuracy gains rather than trading against them.

\begin{figure}[t]
\centering
\includegraphics[height=4.5cm]{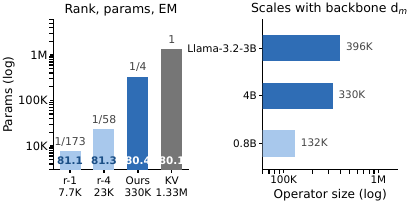}\hspace{0.03\linewidth}%
\includegraphics[height=4.5cm]{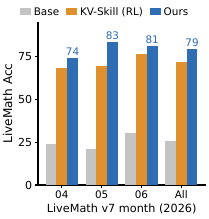}
\caption{\textbf{Left:} the rank--parameter--score ladder on SearchQA: 7.7K parameters already exceed the published 1.33M RL result, and the operator size scales linearly with the backbone width, so smaller models take smaller skills. \textbf{Right:} out-of-distribution generalization. A bundle trained only on problems released through February 2026, evaluated zero-shot on new LiveMath problems from April--June 2026 (zero id overlap): 79.0 versus base 25.4, and 7.5 points above the official released \kvs{} RL artifact under the identical evaluation protocol.}
\label{fig:param_ood}
\end{figure}

\textbf{Out-of-distribution generalization.} A skill worth having survives contact with new data. The LiveMath bundle is trained on problems released through February 2026 and evaluated zero-shot on the April, May, and June 2026 releases, with zero id overlap and a different problem-length distribution (Figure~\ref{fig:param_ood}, right). The gain does not decay: 79.0 versus a base of 25.4 (+53.6) merged, with every month individually 50--61 points above its base (April 74.1 versus 24.1, May 83.2 versus 21.1, June 81.0 versus 30.2). The comparison against the officially released \kvs{} RL artifact, imported and verified in our evaluation pipeline, is the least favorable one for us (its RL versus our BC), and \method{} is higher in every month, by 4.3 to 13.7 points (merged 79.3 vs.\ 71.8, +7.5). The RL artifact's failure mode is diagnostic: on the new months it over-predicts two answer labels at rates far from the gold distribution, while \method{}'s prediction distribution stays near uniform, consistent with the identity anchor keeping the operator's solution label-neutral rather than tuned to the training label statistics. What the operator learned is the task, not the training problems: whether a skill needs retraining is decided by whether the task has changed, not by whether the data has.

\textbf{Training cost.} Table~\ref{tab:training_cost} (Appendix) summarizes the asymmetry: every \method{} skill trains in 12--33 minutes of BC on one GPU, while a single GRPO run of the baseline costs 45--77 minutes plus a resident rollout engine. The entire comparison table of skills costs less to produce than one RL baseline.

\section{Experiments II: why it works}
\label{sec:mechanism}

\begin{figure}[t]
\centering
\includegraphics[height=6.0cm]{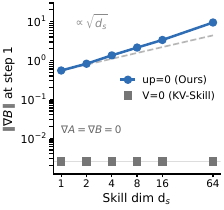}\hspace{0.03\linewidth}%
\includegraphics[height=6.0cm]{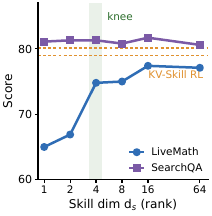}
\caption{\textbf{Left:} gradient-flow probe over the first optimization steps. With $up{=}0$, the output projection $B$ has gradient $\Theta(\sqrt{d_s})$ at step 1; with \kvs{}'s $V{=}0$ anchoring, both large matrices have exactly zero gradient while only the small middle matrix moves. \textbf{Right:} accuracy versus rank, trained from scratch with BC. The knee sits between rank 2 and 4 on LiveMath; SearchQA saturates at rank 1. Rank-4 (23K parameters) reaches the 1.33M level.}
\label{fig:grad_rank}
\end{figure}

\subsection{Why the same BC signal failed before}
\label{sec:trainability}

\kvs{}'s BC control is the natural objection to everything above: the same 400 examples, the same loss, and no learning. The two operators differ, however, in where the identity anchor sits. In \eqref{eq:kvskill} the injected term is proportional to $V$, so at $V{=}0$ the gradient of the loss with respect to both $A$ and $B$ is exactly zero; the only parameters that move in the first steps are the small matrices inside the retrieval bottleneck. In \eqref{eq:decsteer} the anchor is the output projection $B$ itself, so $B$'s first-step gradient is nonzero and grows as $\Theta(\sqrt{d_s})$ with the bottleneck width: measured, $\|\nabla B\|_1$ at step one is 0.55 at $d_s{=}1$ and 9.45 at $d_s{=}64$, while under $V{=}0$ the corresponding gradients are exactly $0$ (Figure~\ref{fig:grad_rank}, left). The optimization path also differs in shape: with $up{=}0$ the down projection $A$ is frozen for two steps while $B$ grows a direction from nothing, and only then aligns to it, reaching the loss of a well-initialized start within 6--10 steps but into a different subspace. Initialization and architecture jointly determine whether an operator trains: the $V{=}0$ anchor leaves both large matrices unreachable at the start, while our combination of a zero-initialized output projection with a shared, normalized, gated backbone keeps them on the optimization path. With signal and data held fixed, the outcomes separate, 68.4 versus 81.1 on SearchQA and 20.2 versus 80.9 on LiveMath, and the same combination reaches the published RL level with 1/58 of the parameters (Section~\ref{sec:rank}). The published control was not short of signal; its large matrices never entered the optimization.

\subsection{How much rank does a decision need?}
\label{sec:rank}

Figure~\ref{fig:grad_rank} (right) sweeps the bottleneck from rank 1 to 64, trained from scratch with the same BC recipe. On LiveMath the curve has a sharp knee between rank 2 and 4 (+7.9 points from rank 2 to rank 4) and is flat beyond; on SearchQA rank 1 saturates. Two controls locate the cause of the knee in optimization rather than capacity. First, the rank-2 model fits the training set to zero loss at every rank, and adding 51\% more data does not move the knee. Second, truncating a trained rank-4 solution to rank 2 loses 4.8 points while training rank 2 from scratch is worse still, and the SVD effective rank of a trained 330K solution is 33--63 with top-1 energy of only 0.03--0.46: the rank-4 solution is not a compressed version of the high-rank one but a different solution path that low-rank training can reach and post-hoc truncation cannot. For practice the numbers are the message: a decision operator needs 23K parameters (1/58 of \kvs{}), and skills whose target is a single forward mapping, such as extraction-style QA, run at 7.7K (1/173).

\subsection{The gain is injection, not compression}
\label{sec:attribution}

\begin{figure}[t]
\centering
\includegraphics[height=4.4cm]{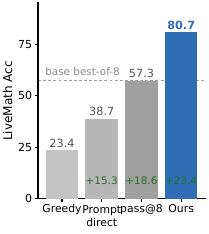}\hspace{0.03\linewidth}%
\includegraphics[height=4.4cm]{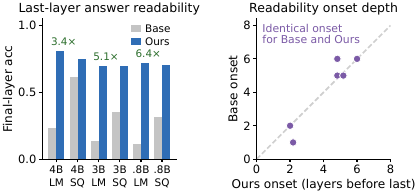}
\caption{\textbf{Left:} decomposing the LiveMath gain. Base greedy 23.4; forced direct answering via prompt 38.7; base best-of-8 sampling 57.3; \method{} 80.7. The last step exceeds the base model's sampling budget: on 43 problems the operator is correct while all 8 base samples are wrong. \textbf{Right:} logit-lens readout of answer readability by layer. The answer becomes readable only within the last 1--6 layers in both arms; the operator amplifies the existing late-layer signal (3.4--6.4$\times$, more for weaker bases) rather than writing the answer earlier.}
\label{fig:attribution}
\end{figure}

Because the operator shortens outputs, the natural suspicion is that it merely suppresses noisy reasoning. Three controls say otherwise (Figure~\ref{fig:attribution}, left). Forcing the base model to answer directly with a prompt recovers 15.3 of the 57 points, which bounds everything that ``stopping meandering'' can contribute. Best-of-8 sampling from the base reaches 57.3, which bounds what the model knows but greedy decoding misses; errors on this task are systematic, since 8 samples recover far less than independent sampling would predict. The operator reaches 80.7, and on 43 problems it is correct while all 8 base samples are wrong: at least 23 points of the gain are answers outside the base model's reachable distribution, injected rather than selected. On CSQA, the same prompt control changes nothing (64/64 symmetric flips), so the direct-answer mode itself is content-neutral there; the accuracy story is carried by the injection, which matches what the headroom trend of Section~\ref{sec:when} leads us to expect for a saturated base.

\subsection{Where the answer is written}
\label{sec:logitlens}

A logit-lens readout of the residual stream localizes the mechanism (Figure~\ref{fig:attribution}, right). Answer information is unreadable through the first two thirds of the network in both the base and steered models, and becomes readable only within the last 1--6 layers; the layer at which readability begins is the same in both arms in 6 of 6 backbone--task combinations (within one layer in the rest). The operator does not inject the answer early. It amplifies the model's own late-layer answer direction, by 3.4$\times$ on the 4B base, 5.1$\times$ on Llama-3.2-3B, and 6.4$\times$ on the 0.8B base, and the amplification grows exactly as the base weakens. This is the representational face of the headroom trend: a stronger base leaves less probability mass for the operator to redistribute, and a late-layer amplifier of fixed training cost scales its effect with the base's uncertainty. It also explains the engineering observations: injecting at 8 depths works as well as 4 or 32 because the effect funnels into the same late layers, and SearchQA calibration improves alongside accuracy (Appendix~\ref{app:results}).

\section{Experiments III: when to steer}
\label{sec:when}

\begin{figure}[t]
\centering
\includegraphics[height=4.4cm]{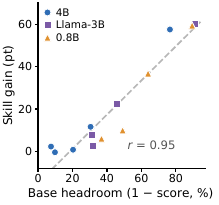}\hspace{0.03\linewidth}%
\includegraphics[height=4.4cm]{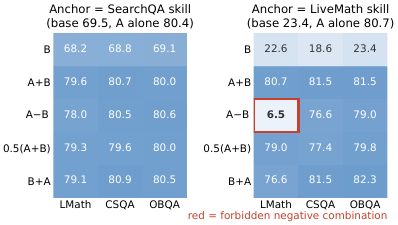}
\caption{\textbf{Left:} the headroom trend, an empirical relation. Skill gain versus base headroom ($1-$ base score) over 13 base--task pairs on three backbones; the relation is monotone, and saturated bases (CSQA, OpenBookQA on 4B) sit at zero as the trend expects. \textbf{Right:} skill algebra. Adding, interpolating, and reordering two skills preserves the anchor skill in all 18 cells (zero interference); the one forbidden zone is a negative coefficient on a mode-switching skill, which reverses the modality it switches.}
\label{fig:when}
\end{figure}

\subsection{An empirical headroom trend}
\label{sec:headroom}

Across every base--task pair we measured, the gain of a trained operator tracks how much room the base has left: plotting gain against $1 - \text{base score}$ gives a consistently increasing relation over the 13 base--task pairs, spanning Qwen3.5-4B, Llama-3.2-3B, and Qwen3.5-0.8B (Figure~\ref{fig:when}, left). We treat this as an empirical regularity and a screening heuristic rather than a law: the 13 points are not independent, since several tasks share a backbone, and we do not lean on a correlation coefficient; what matters in practice is that the relation held at every pair we measured, and it earns its keep on the pairs that look like failures. On the 4B base, CSQA and OpenBookQA are saturated (79.7 and 90.2), and there all 24 skill arms we trained, across every operator variant and training signal, land within about a point of the base, with paired flip counts almost perfectly symmetric: the operator changes \emph{which} problems are answered correctly but not \emph{how many}. Read through this trend, this is not a limitation of the operator but a measurement of the base, and the trend suggests a testable prediction: weaken the base and the same operator, the same BC signal, and the same hyperparameters should produce gains. They do, on the same tasks the 4B base saturates: +2.4 to +10.0 points on CSQA and OpenBookQA for Llama-3.2-3B and Qwen3.5-0.8B. Headroom is therefore the first screening criterion for where to spend a skill: measure the base, and skip the 12 minutes of training when the base has nothing left to give.

\subsection{An operator algebra for skill libraries}
\label{sec:algebra}

Because each skill is an additive perturbation of the same residual stream, two skills can be applied together, and the composition behaves like an algebra rather than an arbitrary mixture. Across all 18 cells of our combination study (three partner skills, three strengths, two anchor tasks), adding, half-interpolating, and reordering skills preserves the anchor skill's full gain, never falls below the base, and produces interaction terms far smaller than the main effects; the order of application, although not commutative in the vector norm, changes decisions by at most 4 points. One boundary is sharp: applying a negative coefficient to a mode-switching skill (one whose job is to flip long reasoning into direct answering) reverses the switch and collapses accuracy below the base, and lowering the strength does not recover it. The practical contract is simple: skills can be added, interpolated, reordered, and hot-swapped per request with positive coefficients; subtraction is reserved for skills that shift content, not skills that switch modes.

\subsection{Strength as an inference-time knob}
\label{sec:strength}

The scalar $\alpha$ in \eqref{eq:decsteer} exposes skill strength without retraining. Sweeping it across four tasks with 3 seeds each shows clean modulation with mild overshoot at full strength on two tasks (LiveMath prefers 0.75, CSQA 0.5, each by about one point over scale 1.0) and no overshoot on the other two (Appendix~\ref{app:setup}). All conclusions in this paper hold under a uniform scale of 1.0 (Table~\ref{tab:main_results} caption), so the knob is a deployment control rather than a hidden tuning axis.

\section{Related work}
\label{sec:related}

\textbf{Externalizing skills.} A growing body of work moves competence out of weights and into portable objects: textual skill files and libraries for agents \citep{tang2026knowledge,alzubi2026evoskill,wang2026skillx,wang2026reinforcement,ling2026agent,liu2026agent}, trajectory-level distillation into reusable lessons \citep{ni2026trace2skilldistilltrajectorylocallessons,lu2026skill0}, and compressed behavioral soft prompts \citep{bai2026skillzip,li2026skillsinjector,yang2026skillopt,tao2026softskill}. \citet{han2026kv} is the closest work and our reference point: it forges a skill directly in the model's native language as a factorized residual-stream operator trained with RL. We keep its injection site and identity-anchor discipline but show that the operator can be two orders of magnitude smaller, that BC succeeds once the anchor and the architecture keep the large matrices on the optimization path, and that the resulting objects compose algebraically.

\textbf{Activation steering and parameter-efficient adaptation.} Additive interventions on hidden states, from activation engineering \citep{turner2023steering,subramani2022extracting} and inference-time interventions \citep{li2023inference} to representation finetuning \citep{wu2024reft}, function vectors \citep{todd2024function}, and principled frameworks that treat weight updates as activation shifts \citep{adila2026weightupdatesactivationshifts}, establish that small directional edits can control model behavior; related work analyzes their reliability \citep{tan2024analysing}, scaling \citep{stoehr2024activation}, subspace structure \citep{zhao2025beyond}, and safety applications \citep{zou2024improving,park2025steer}. These methods typically steer concepts or refuse behaviors; \method{} steers \emph{task decisions}, trains against answer tokens with a task-level verifier, and contributes a trainability analysis of the identity anchor that we have not seen articulated for this operator family. Continuous-prompt methods \citep{li2021prefix,lester2021power,liu2022p} and automatic differentiation over text \citep{yuksekgonul2024textgrad} share the small-parameter goal but act through the input or the prompt rather than the residual stream.

\textbf{Distilling System 2 into System 1.} Compiling deliberation into intuition has been pursued at the data level, by generating and distilling reasoning traces \citep{yu2024distilling,zelikman2024quietstarlanguagemodelsteach,liao2025longperceptualthoughts,yang2025llm2}, at the architecture level, with latent reasoning slots and randomized traces \citep{hao2024training,su2025dualformer,weston2023system}, and at the policy level, by learning \emph{when} to think \citep{fang2026thinkless,yong2026think,ziabari2025reasoning}. \method{} sits at a different point in this design space: nothing is distilled into weights, and no routing governs when to think. A frozen model preserves its full System-2 capability, while a 23K-parameter side operator carries the compiled decision , keeping skills swappable, composable, and instantly removable.

\section{Limitations and conclusion}
\label{sec:limitations}

\textbf{Limitations.} The operator's natural habitat is tasks whose target behavior is a single forward mapping, such as extraction-style QA, direct answering, and modality switching; on tasks where the chain of reasoning itself carries information the gain is governed by the headroom trend rather than by recovered reasoning, although the base model can still be prompted for deliberation at any time. Operator parameters bind to the backbone width and depth, so transferring a skill across model sizes means retraining on the target, at a cost of minutes. On saturated bases the operator cannot, and should not, be expected to add accuracy; its value there is the inference speedup. Negative coefficients are safe for content skills but not for mode-switching skills. Finally, our strongest comparisons quote published \kvs{} numbers under their protocol; the fully controlled same-protocol comparisons in Appendix~\ref{app:results} cover the smaller backbones and the released RL artifact.

\textbf{Conclusion.} We set out to price a skill and found the price is far lower than the current standard: a 23K-parameter residual operator, twelve minutes of behavior cloning, and a 6-token output reproduce what previously took 1.33M parameters and a reinforcement-learning pipeline, on five tasks and three backbones, with the gain persisting out of distribution. The mechanism is now legible: identity anchoring at the output projection keeps the large matrices trainable, the operator decides by amplifying the base model's own late-layer answer pathway, and the headroom trend indicates where this amplification will pay. A skill that costs minutes to forge and swaps like a cartridge changes what a skill library can be, and rank-4 suffices to decide on SearchQA and near-suffices on LiveMath, where higher rank still helps.

\subsubsection*{Reproducibility statement}
All experiments use public models and datasets with the splits, prompts, decoding, and evaluators of \citet{han2026kv}. Operator architectures, initialization, training hyperparameters, seeds, identity-anchor checks, contamination screens, and the strength-sweep protocol are specified in Appendix~\ref{app:setup} and Appendix~\ref{app:rank}; training and evaluation code is included in the supplementary material.

\bibliography{refs}
\bibliographystyle{iclr2027_conference}

\appendix

\section{Experimental setup}
\label{app:setup}

\textbf{Models and tasks.} Main backbone: Qwen3.5-4B ($d_m{=}2560$, 32 blocks, hybrid attention). Cross-backbone: Llama-3.2-3B-Instruct (28 blocks, $d_m{=}3072$) and Qwen3.5-0.8B (24 blocks, $d_m{=}1024$). Tasks: LiveMath (accuracy, $n{=}124$; plus the 2026-04/05/06 releases, $n{=}319$), SearchQA (EM, $n{=}1400$), CommonsenseQA (accuracy, $n{=}1221$), OpenBookQA (accuracy, $n{=}500$), DocVQA (ANLS, $n{=}374$). Decoding is greedy with temperature 0, stop string \texttt{</answer>}, budgets 1024 for multiple choice, 8192 for LiveMath, and at least 4096 for retrieval-style tasks; thinking mode is disabled in prompts. Metrics and splits follow \kvs{} exactly, using the same split files and each task's native evaluator.

\textbf{Protocol checks.} Table~\ref{tab:protocol_checks} summarizes the integrity checks behind the reported numbers: identity anchors at $\theta^0$ and at scale 0 are verified item-by-item against the bundle-free base; LiveMath train--test overlap is zero; gold label distributions are near uniform with majority-class baselines near 25\%, and prediction distributions show no collapse; the official released \kvs{} RL artifact, imported into our evaluation pipeline, reproduces its self-reported score to within one problem (78.2 vs.\ 79.0).

\begin{table}[t]
\centering
\small
\setlength{\tabcolsep}{4.5pt}
\caption{Protocol and integrity checks behind the reported numbers: exact
	identity anchors for fair comparison, leakage and degeneracy screens, budget
	audits, and the import validation of the official \kvs{} artifact used as a
	baseline.}
\label{tab:protocol_checks}
\begin{tabular}{l p{8.5cm}}
\toprule
Check & Result \\
\midrule
Identity anchor ($\theta^0$, up$=0$) & Output bit-identical to base (max$|{\Delta}|\le 0.0625$, bf16 rounding) \\
Identity at scale $=0$ (n=400 / n=50) & Item-identical to the no-bundle base \\
Train $\cap$ test overlap (LiveMath) & 0 duplicated ids \\
LiveMath v7 vs.\ train split & 0 / 319 id overlap (2026-04/05/06) \\
Gold label distribution (LiveMath) & $\approx$uniform; majority class 28.2\% \\
Prediction distribution (LiveMath) & $\approx$uniform; no single-label collapse \\
Budget truncation & MCQ 1024 / LiveMath 8192 / STaRK $\ge$4096; v7 verified no silent truncation at 16384 \\
\kvs{} artifact import & Reproduces self-score within 1 item (78.2 vs.\ 79.0) \\
Decoding consistency (6-token arm) & Monthly scores agree within 1 item across evaluators \\
\bottomrule
\end{tabular}
\end{table}

\begin{figure}[h]
\centering
\includegraphics[width=0.9\linewidth]{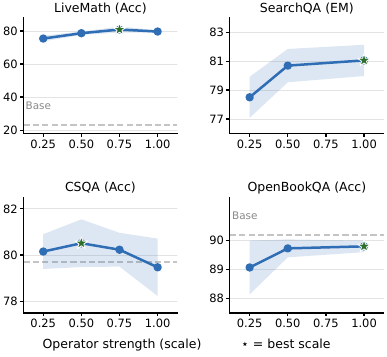}
\caption{Operator strength sweeps (mean $\pm$ SD over seeds). Stars mark each task's best strength.}
\label{fig:scale}
\end{figure}

\textbf{Operator strength protocol.} The scalar $\alpha$ (scale) modulates skill strength at inference. Table-scale sweeps with 3 seeds per task (Appendix Figure~\ref{fig:scale}) show clean, monotone modulation on all tasks with mild overshoot at full strength on LiveMath (best at 0.75, +1.2 points over 1.0) and CSQA (best at 0.5, +1.0), and none on SearchQA and OpenBookQA. The main table uses each dataset's best strength; a uniform scale of 1.0 changes no conclusion (LiveMath 79.6, SearchQA 80.9, CSQA 79.5, OpenBookQA 89.8, DocVQA 93.2).

\textbf{Main-table conventions.} LoRA rows quote the \kvs{} paper's LoRA baseline (1.38M parameters). In the cross-backbone blocks no other published baselines exist; the \kvs{} operator there has 0.54--1.59M parameters against 0.13--0.40M for ours, and the 0.8B LiveMath base was not measured. The rank-scan rows were scanned on LiveMath and SearchQA only.

\textbf{Seeds and artifact integrity.} Key arms use 3 training seeds with mean $\pm$ SD reported. Run-to-run variance on small-data tasks is dominated by bf16 drift after the first steps, so conclusions are drawn from multi-seed aggregates rather than single runs; artifact-integrity screening is part of our seeding protocol.

\textbf{Training details.} BC uses Adam with learning rate $10^{-3}$, batch size 1 with accumulation 4, loss on answer tokens only. LiveMath: 35 examples, 10 epochs (87 steps, 12 min). SearchQA: 400 examples, 2 epochs (200 steps, 33 min). DocVQA: 107 examples, 78 steps (29 min, gradient checkpointing). GRPO comparisons follow \kvs{} hyperparameters exactly (300 steps, 8 prompts $\times$ 8 rollouts, $T{=}1.0$, $\epsilon_{\text{low}}{=}0.2$, $\epsilon_{\text{high}}{=}0.28$), with the one deviation documented below and in Appendix~\ref{app:rl}: the published entropy bonus ($\beta_{\text{ent}}{=}0.05$) overwhelms the KL anchor by two orders of magnitude on long-CoT tasks and diverges; setting $\beta_{\text{ent}}{=}0$ with $\beta_{\text{kl}}{=}0.03$ is stable and is used for all fixed-recipe runs. Table~\ref{tab:grpo_multiseed} gives the multi-seed comparison under the identical protocol.

\section{Additional results}
\label{app:results}

\begin{table}[t]
\centering
\setlength{\tabcolsep}{10pt}
\caption{Behavior cloning matches GRPO under \kvs{}'s own protocol, at 1/4 the
	parameters. All BC seeds meet or exceed the \kvs{} RL reference (80.1 / 79.0).
	On LiveMath, GRPO from scratch trails both its \kvs{} counterpart and the BC
	warm start: the starting point matters more than the algorithm.}
\label{tab:grpo_multiseed}
\begin{tabular}{llccc}
\toprule
Task & Method & Seeds & Score $\pm$ SD & \kvs{} \\
\midrule
SearchQA & Ours (BC)              & 3 & \textbf{80.9 $\pm$ 0.6} & 80.1 \\
SearchQA & Ours (GRPO scratch)    & 3 & 80.7 $\pm$ 0.9 & 80.1 \\
SearchQA & Ours (GRPO from skill) & 1 & 80.6 & 80.1 \\
LiveMath & Ours (BC)              & 4 & \textbf{79.6 $\pm$ 0.8} & 79.0 $\pm$ 3.9 \\
LiveMath & Ours (GRPO from skill) & 1 & 80.7 & 79.0 $\pm$ 3.9 \\
LiveMath & Ours (GRPO scratch)    & 1 & 71.8 & 75.6 $\pm$ 4.1 \\
\bottomrule
\end{tabular}
\end{table}

\textbf{Same-protocol RL comparison.} Table~\ref{tab:grpo_multiseed}: under GRPO identical to \kvs{}'s main protocol, \method{} matches the published numbers with 1/4.02 parameters, and RL does not exceed BC on either task with headroom. GRPO from scratch on LiveMath (71.8 vs.\ 80.7 from the BC warm start) shows that the starting point matters more than the algorithm, the same conclusion the trainability analysis reaches from the gradient side.

\begin{table}[t]
\centering
\setlength{\tabcolsep}{6pt}
\caption{Cross-carrier comparison: the \kvs{} operator is reimplemented exactly
	($V{=}0$ identity init) and both operators are trained and evaluated under
	identical data, prompts, and decoding. Ours wins 6 of 8 cells and is never
	worse, with 1/4 the parameters.}
\label{tab:cross_carrier}
\begin{tabular}{llcccc}
\toprule
Model & Task & Base & Ours (3 seeds) & \kvs{} (3 seeds) & $\Delta$ \\
\midrule
Llama-3.2-3B  & SearchQA & 57.1 & \textbf{77.6 $\pm$ 1.3} & 76.5 $\pm$ 0.1 & $+1.1$ \\
Llama-3.2-3B  & LiveMath & 16.9 & \textbf{71.2 $\pm$ 2.0} & 66.1 $\pm$ 1.4 & $+5.1$ \\
Llama-3.2-3B  & CSQA     & 71.3 & \textbf{72.7 $\pm$ 0.9} & 71.7 $\pm$ 0.6 & $+1.0$ \\
Llama-3.2-3B  & OpenBookQA & 74.6 & 78.2 $\pm$ 0.7       & \textbf{78.9 $\pm$ 0.8} & $-0.7$ \\
Qwen3.5-0.8B  & SearchQA & 36.1 & \textbf{72.9 $\pm$ 0.8} & 70.1 $\pm$ 0.7 & $+2.8$ \\
Qwen3.5-0.8B  & LiveMath & --   & \textbf{70.7 $\pm$ 3.3} & 62.6 $\pm$ 3.1 & $+8.1$ \\
Qwen3.5-0.8B  & CSQA     & 50.9 & \textbf{61.1 $\pm$ 1.1} & 61.0 $\pm$ 1.3 & $+0.1$ \\
Qwen3.5-0.8B  & OpenBookQA & 63.2 & \textbf{69.8 $\pm$ 1.2} & 66.7 $\pm$ 0.6 & $+3.1$ \\
\bottomrule
\end{tabular}
\end{table}

\textbf{Cross-carrier comparison.} Table~\ref{tab:cross_carrier} reruns the full comparison with the \kvs{} operator faithfully reimplemented (including $V{=}0$ anchoring) with identical data, prompts, and decoding: \method{} is higher in 6 of 8 cells and never lower, at 0.13--0.40M versus 0.54--1.59M parameters.

\begin{table}[t]
\centering

\setlength{\tabcolsep}{6pt}
\caption{Training cost on a single A100-40GB. The BC recipe used for all main
	results trains in minutes because there are no rollouts; GRPO rows are the
	protocol-matched controls of Table~\ref{tab:grpo_multiseed}. Inference
	additionally accelerates generation by up to 36$\times$ (LiveMath), 10.3$\times$
	wall-clock (DocVQA).}
\label{tab:training_cost}
\begin{tabular}{llrrr}
\toprule
Task & Signal & Examples & Wall-clock & Peak mem. \\
\midrule
LiveMath (4B)  & BC    & 35  & 12 min & 13.9 GiB \\
SearchQA (4B)  & BC    & 400 & 33 min & -- \\
DocVQA (4B)    & BC    & 107 & 29 min & 27.9 GiB \\
CSQA (4B)      & RFT   & 400 & --     & -- \\
SearchQA (4B)  & GRPO  & 300 steps & 45 min & 12.6 GiB \\
LiveMath (4B)  & GRPO  & 300 steps & 77 min & 13.6 GiB \\
CSQA (4B)      & GRPO  & 300 steps & 20 min & 13.1 GiB \\
\bottomrule
\end{tabular}
\end{table}

\begin{figure}[h]
\centering
\includegraphics[width=0.9\linewidth]{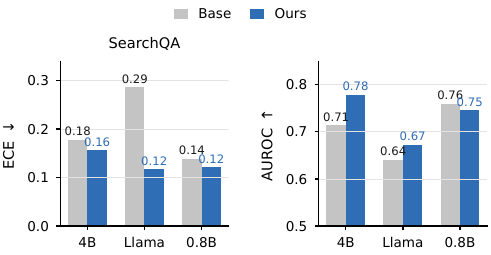}
\caption{SearchQA calibration before and after skill injection: ECE (left) and answer-ranking AUROC (right) across the three backbones.}
\label{fig:calibration}
\end{figure}

\textbf{Confidence and calibration.} Figure~\ref{fig:calibration} reports SearchQA calibration before and after injection across the three backbones. ECE drops on every backbone (0.29 to 0.12 on Llama, 0.18 to 0.16 on 4B, 0.14 to 0.12 on 0.8B), and AUROC improves on 4B and Llama (0.71 to 0.78, 0.64 to 0.67) while staying flat on 0.8B. We restrict the comparison to SearchQA: on LiveMath the 6-token mode saturates answer confidence for both methods, which makes calibration readouts uninformative there; Appendix~\ref{app:diagnostics} discusses why internal confidence signals are limited in general.

\begin{figure}[h]
\centering
\includegraphics[height=6cm]{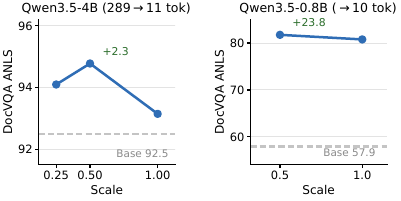}\hspace{0.03\linewidth}%
\includegraphics[height=6cm]{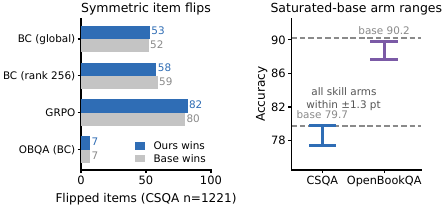}
\caption{\textbf{Left:} multimodal extension on DocVQA. The identical BC recipe transfers to document VQA: 4B gains +2.3 ANLS (92.5 to 94.8, higher in 3/3 seeds) and 0.8B gains +23.8 (57.9 to 81.8), with a second confirmation of mild overshoot at full strength. \textbf{Right:} behavior on saturated 4B bases. Paired flip counts are almost perfectly symmetric (53/52, 58/59, 82/80, 7/7) and all skill arms sit within about a point of the base, the headroom trend's expectation for a saturated base.}
\label{appfig:docvqa_sat}
\end{figure}

\section{RL stability: the entropy-bonus divergence}
\label{app:rl}

\begin{figure}[h]
\centering
\includegraphics[height=6cm]{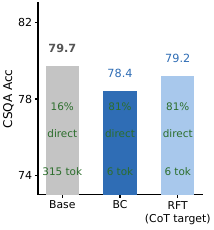}\hspace{0.03\linewidth}%
\includegraphics[height=6cm]{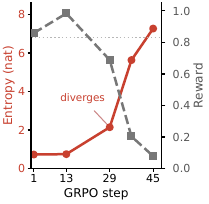}
\caption{\textbf{Left:} RFT control on CSQA. Even when the training target contains full chains of thought (330/400 sampled solutions kept by the verifier, median target 1,445 characters), the operator still converges to 6-token direct answers with accuracy unchanged: the direct-answer mode is the operator's least-effort solution, not an artifact of short targets. \textbf{Right:} RL stability. The published entropy bonus overwhelms the KL anchor by roughly 100$\times$ once entropy rises, driving divergence (reward 0.86 to 0.08); removing it restores stability. This supports using BC, which needs none of these controls.}
\label{appfig:rft_rl}
\end{figure}

\textbf{RFT control and the entropy divergence.} Two auxiliary findings complete the training-signal story (Figure~\ref{appfig:rft_rl}). The RFT control shows that the direct-answer collapse is structural rather than an artifact of short targets, and the GRPO trace shows why the RL baseline needs careful recipe control that BC avoids entirely.

\section{Rank scan and trainability details}
\label{app:rank}

\textbf{Full rank tables.} Five-seed from-scratch results (LiveMath accuracy): rank 1: 65.0, rank 2: 66.9, rank 4: 74.8, rank 8: 75.0, rank 16: 77.4, rank 64: 77.1. SearchQA EM: 81.1, 81.3, 81.3, 80.8, 81.7, 80.6. The knee on LiveMath is between ranks 2 and 4 (+7.9 points); SearchQA saturates at rank 1.

\textbf{Capacity versus optimization.} Three controls separate the two hypotheses. (i) All ranks fit the training set to zero loss, and raising LiveMath from 35 to 53 training examples leaves the knee unchanged, so data and fit are not the bottleneck. (ii) Gradient statistics grow smoothly with $d_s$ under $up{=}0$ with no inflection at the knee, so gradient flow alone does not predict it either. (iii) Truncation is lossy and path-dependent: a trained rank-4 solution truncated to rank 2 drops 4.8 points zero-shot, truncation to rank 1 drops to 42.7 (base is 23.4), and whether truncation helps or hurts relative to from-scratch training depends on the source rank. The consistent picture is that each rank converges into its own subspace (SVD effective rank of a trained 330K operator is 33--63 with top-1 energy 0.03--0.46), so low-rank solutions are alternative solution paths, not compressions of high-rank ones.

\section{Diagnostic probes}
\label{app:diagnostics}

\begin{figure}[h]
\centering
\includegraphics[width=0.6\linewidth]{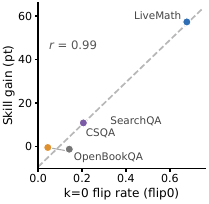}
\caption{Commitment-point probe: applying a suffix intervention from generation step $k$ measures how early the answer is fixed. The height of the curve at $k{=}0$ tracks the eventual skill gain across the four tasks probed and follows the headroom trend, so it serves as an auxiliary screening diagnostic; the informative quantity is the curve's height, not its onset.}
\label{appfig:commitment}
\end{figure}

\textbf{Commitment-point probe.} A single-point or suffix intervention applied from generation step $k$ measures how early the answer is committed (Figure~\ref{appfig:commitment}). Curve height at $k{=}0$ separates injectable tasks ($\geq 0.20$: LiveMath 0.68, SearchQA 0.21) from saturated ones ($\leq 0.15$) across the four tasks we probed; because it is collinear with the cheaper headroom statistic, we use it as a secondary screening heuristic only.

\textbf{Gating statistics are not per-sample confidence.} We tested whether internal gate activations could serve as per-sample confidence for selective compression. They cannot: gate statistics vary by only 1.7--3.6\% across samples and are collinear with prompt length ($\rho$ up to 0.9), so any apparent correlation with compression gain is a length effect. Internal skill confidence, if needed, should come from a trained head or external signals such as self-consistency.

\section{LLM usage }
\label{app:llm}
Large-language-model tools assisted with editing and formatting during manuscript preparation. All research questions, experimental design, runs, analysis, and claims are the authors' own; all reported numbers were generated by the experimental pipeline described above.

\end{document}